\documentclass[conference]{IEEEtran}
\IEEEoverridecommandlockouts

\usepackage{cite}
\usepackage{textcomp}
\usepackage{amsfonts}

\makeatletter
\def\input@path{table/}
\makeatother

\usepackage{caption}
\usepackage{subcaption}
\usepackage{graphicx}
\usepackage{amsthm}
\usepackage{algorithm}
\usepackage{algorithmic}
\usepackage{enumitem}
\usepackage{amsmath}
\usepackage{amssymb}
\usepackage{xcolor}
\usepackage{booktabs}
\usepackage{multirow}
\usepackage{url}

\theoremstyle{definition}

\usepackage[hidelinks]{hyperref}

\graphicspath{image/}

\newcommand{\modelname}{GAD-MoRE}

\begin{document}

\title{Zero-shot Generalizable Graph Anomaly Detection with Mixture of Riemannian Experts}

\author{
Xinyu Zhao$^{1}$, Qingyun Sun$^{1\dagger}$, Jiayi Luo$^{1}$,
Xingcheng Fu$^{2}$, Jianxin Li$^{1}$ \\
$^{1}$\textit{School of Computer Science and Engineering, Beihang University, Beijing, China} \\
$^{2}$\textit{Key Lab of Education Blockchain and Intelligent Technology,Ministry of Education, }\\
\textit{Guangxi Normal University, Guilin, China} \\
\{xyzhao, sunqy, luojy, lijx\}@buaa.edu.cn, fuxc@gxnu.edu.cn
}

\maketitle

\begingroup
\renewcommand\thefootnote{\fnsymbol{footnote}}
\footnotetext[2]{Corresponding author.}
\endgroup

\begin{abstract}

Graph Anomaly Detection (GAD) aims to identify irregular patterns in graph data, and recent works have explored zero-shot generalist GAD to enable generalization to unseen graph datasets.
However, existing zero-shot GAD methods largely ignore intrinsic geometric differences across diverse anomaly patterns, substantially limiting their cross-domain generalization.
In this work, we reveal that anomaly detectability is highly dependent on the underlying geometric properties and that embedding graphs from different domains into a single static curvature space can distort the structural signatures of anomalies.
To address the challenge that a single curvature space cannot capture geometry-dependent graph anomaly patterns, we propose \underline{\modelname}, a novel framework for zero-shot Generalizable \underline{G}raph \underline{A}nomaly \underline{D}etection with a \underline{M}ixture \underline{o}f \underline{R}iemannian \underline{E}xperts architecture.
Specifically, to ensure that each anomaly pattern is modeled in the Riemannian space where it is most detectable, \modelname~employs a set of specialized Riemannian expert networks, each operating in a distinct curvature space.
To construct topology-aware inputs for the subsequent Riemannian experts, we introduce an anomaly-aware multi-curvature feature alignment module that combines dimensionality reduction with Laplacian feature selection over parallel feature branches.
Finally, to facilitate better generalization beyond seen patterns, we design a memory-based dynamic router that adaptively assigns each input to the most compatible expert based on historical reconstruction performance on similar anomalies.
Extensive experiments in the zero-shot setting demonstrate that \modelname{} significantly outperforms state-of-the-art generalist GAD baselines\footnote{Our code is available at \url{https://github.com/zhaoxinyu2002/GAD-MoRE}.}.
\end{abstract}


\section{Introduction}
\label{sec:intro}

Graph Anomaly Detection (GAD)~\cite{ma2021comprehensive,qiao2025deep,yuan2025comprehensive}
aims to identify irregular patterns in graph-structured data.
Detecting such anomalies is important in many real-world applications, 
where abnormal nodes may correspond to financial fraud rings~\cite{wang2019semi} 
or fraudulent users in review networks~\cite{rayana2015collective,kumar2018rev2}, 
leading to substantial financial losses and negative social consequences.
Benefiting from the powerful representation learning capability of Graph Neural Networks (GNNs), numerous GAD methods incorporate GNNs into their frameworks~\cite{nassif2021machine, qiao2025deep}.
However, the majority of existing methods are designed under a restrictive assumption: training and inference are performed on the same graph.
This paradigm inherently limits practical applicability, where the structural and attribute patterns that characterize anomalous behavior are not universal across graphs~\cite{ma2021comprehensive}. 
For example, the topology of a spam-bot community in a social network~\cite{yang2019mining} can be entirely different from that of a financial fraud ring in a transaction graph~\cite{wang2019semi}. 
As a result, models trained within a single domain tend to overfit to their specific anomaly patterns, failing to generalize to unseen graphs without costly retraining and labeling. 
This inability to transfer learned anomaly patterns across varying distributions constitutes a critical generalization challenge~\cite{ma2021comprehensive, qiao2025deep}.

\begin{figure*}[t]
  \centering
  \includegraphics[width=\linewidth]{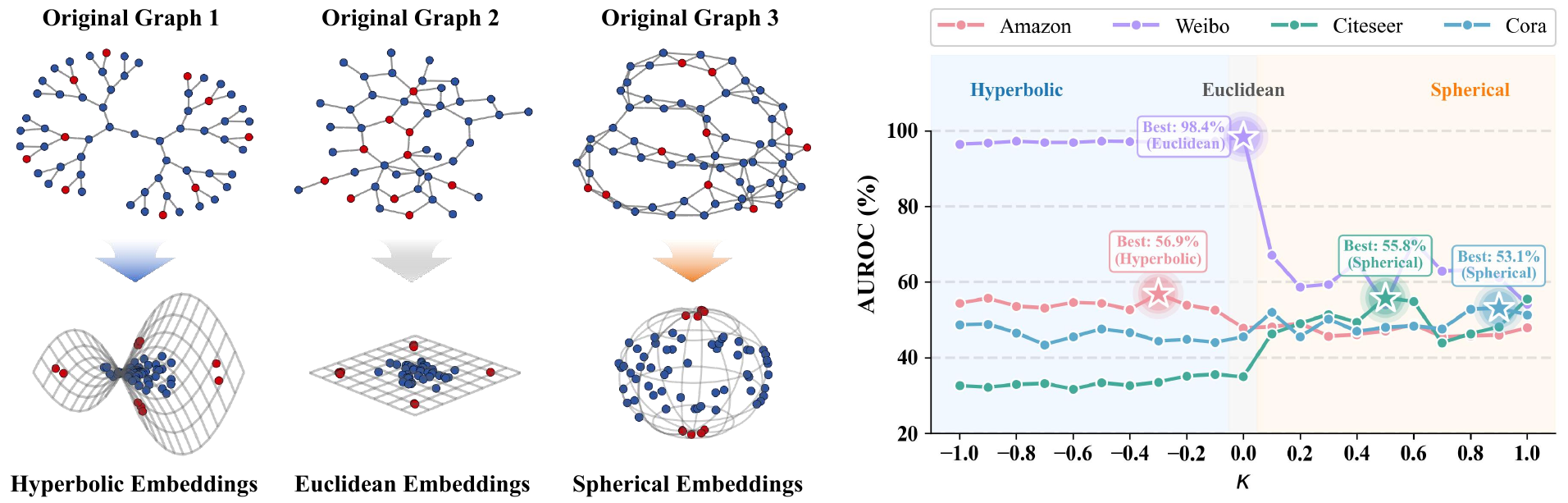} 
    \caption{Geometric sensitivity of graph anomalies. Left: mapping graphs into hyperbolic, Euclidean, and spherical spaces can change anomaly separability. Right: average AUROC of K-Means~\cite{hartigan1979algorithm}, k-NN~\cite{cover1968estimation}, and Isolation Forest~\cite{liu2008isolation} varies with curvature on Amazon, Weibo, Cora, and CiteSeer~\cite{rayana2015collective,kumar2019predicting,sen2008collective}.}
  \label{fig:geometric_sensitivity_experiment}
\end{figure*}

To enable cross-graph generalization,
recent works have begun to explore zero-shot or few-shot GAD models~\cite{liu2024arc,niu2024zero,qiao2025anomalygfm,zhangia}. 
While these approaches have made promising progress, \textit{they often overlook the critical issue that different anomaly patterns exhibit substantial geometric heterogeneity across graphs}. 
As a result, their ability to generalize to new, unseen graph domains remains limited.
Under this heterogeneity, representing all graphs in a single fixed-curvature space can introduce systematic embedding distortion, where certain structural relations are over-compressed or over-stretched, reducing separability and making subtle anomalies harder to distinguish from normal variability.
Geometry acts as an inductive bias that shapes which structural properties are explicitly captured.
For example, in hierarchical graphs, anomalous bridge nodes linking distant branches are naturally exposed as outliers in hyperbolic space with negative curvature~\cite{grover2025curvgad}.
By contrast, graphs with dense structures are better captured in spherical space, where anomalies typically manifest as isolated nodes deviating from clusters~\cite{liu2008isolation,liu2021anomaly}. 
Therefore, imposing a single fixed-curvature across all graphs ignores the geometric heterogeneity of anomaly patterns, potentially distorting anomaly representations.

To examine the geometric heterogeneity of graph anomalies, we conduct a preliminary study on representative datasets, with the results shown in Figure~\ref{fig:geometric_sensitivity_experiment}.
Specifically, we use a standard Graph Autoencoder (GAE)~\cite{kipf2016variational} to learn node embeddings in constant-curvature spaces with different curvature values, and then evaluate anomaly detection performance using classical unsupervised detectors, including K-Means~\cite{hartigan1979algorithm}, k-NN~\cite{cover1968estimation}, and Isolation Forest~\cite{liu2008isolation}. 
As shown in Figure~\ref{fig:geometric_sensitivity_experiment}, performance varies substantially with the choice of curvature.
For example, the Weibo~\cite{kumar2019predicting} dataset shows a clear performance peak in Euclidean space ($\kappa=0.0$), whereas the Citeseer and Cora~\cite{sen2008collective} datasets consistently prefer positive curvature spaces, and the Amazon~\cite{rayana2015collective} dataset exhibits a preference for negative curvature. 
These results motivate that \textit{different anomalies are more effectively detected in different Riemannian spaces, making fixed-curvature embeddings suboptimal for anomaly detection.}

To overcome the limitation that a single fixed Riemannian space cannot capture curvature-dependent anomaly patterns across different graphs, we propose \underline{\modelname}, a novel framework for zero-shot generalizable \underline{G}raph \underline{A}nomaly \underline{D}etection with a \underline{M}ixture \underline{o}f \underline{R}iemannian \underline{E}xperts architecture, explicitly designed to model geometric heterogeneity in graph anomaly patterns.
Specifically, to ensure each anomaly pattern is modeled in the geometry where it is most detectable, we first employ a collection of specialized Riemannian expert networks within the MoE architecture. Unlike single-geometry models that distort structural patterns, each expert operates in a distinct curvature space, enabling distinct anomaly patterns to be captured in their suitable Riemannian space.
Second, to transform raw node features into geometry-sensitive representations that are suitable for these Riemannian experts, we introduce an anomaly-aware multi-curvature feature alignment module. This component bridges the gap between geometry-agnostic attributes and Riemannian manifolds, producing robust representations compatible with diverse experts. 
Finally, to route each node to its most appropriate Riemannian expert, we design a memory-based dynamic router that adaptively assigns current input nodes according to each expert's historical anomaly detection performance, thereby enhancing generalization to unseen data.
Extensive experiments demonstrate that our proposed \modelname{} achieves state-of-the-art performance in challenging zero-shot GAD settings.

Our contributions can be summarized as follows:
\begin{itemize}
    \item We show that graph anomalies exhibit geometric heterogeneity, where different anomaly patterns have different detectability in distinct Riemannian spaces, whereas a single fixed Riemannian space is inherently insufficient to capture them.
    
    \item We propose a novel zero-shot generalizable graph anomaly detection framework named \modelname, which explicitly models graph anomaly geometric heterogeneity via a mixture of Riemannian experts coupled with an anomaly-aware multi-curvature feature alignment module and a memory-based dynamic router.
    
    \item Extensive experiments demonstrate that \modelname{} significantly outperforms state-of-the-art baselines in challenging zero-shot settings.
\end{itemize}

\section{Related Work}
\label{sec:related}

\subsection{Graph Anomaly Detection}
Node-level graph anomaly detection (GAD) aims to identify abnormal nodes in graph-structured data~\cite{ma2021comprehensive,qiao2025deep}. Existing methods mainly include supervised GNN-based detectors~\cite{dou2020enhancing,gao2023addressing} and unsupervised detectors based on reconstruction, contrastive learning, knowledge distillation, adversarial training, or anomaly score modeling~\cite{ding2019deep,fan2020anomalydae,chen2020generative,yuan2021higher,velivckovic2018deep,liu2021anomaly,lin2023discriminative,tian2023sad}. Although effective on individual graphs, these methods often depend on graph-specific supervision or proxy objectives and thus struggle to transfer across domains with different structures and anomaly patterns. Recent generalizable GAD methods, including ARC, UNPrompt, AnomalyGFM, and IA-GGAD~\cite{liu2024arc,niu2024zero,qiao2025anomalygfm,zhangia}, explore few-shot or zero-shot transfer across graphs. However, they mainly pursue domain-invariant representations within a single geometric view, leaving the intrinsic geometric heterogeneity of graph anomalies underexplored.

\subsection{Mixture of Experts for Graph Learning}
Mixture-of-Experts (MoE) models route inputs to specialized experts, improving capacity and conditional computation~\cite{jacobs1991adaptive,shazeer2017outrageously,fedus2022switch}. Recent graph MoE studies use expert routing to handle structural heterogeneity, data imbalance, modality differences, or curvature-aware representations~\cite{hu2021graph,zeng2023mixture,wang2023graph,guo2025graphmore}. In anomaly detection, MoEGAD~\cite{cai2025moegad} explores MoE for graph-level anomaly detection. These works mainly target supervised or single-domain settings, whereas zero-shot generalizable node-level GAD requires expert selection that remains reliable under cross-graph distribution shifts. Our framework addresses this challenge with Riemannian experts and a memory-based dynamic router.

\subsection{Riemannian Graph Learning}
Riemannian graph learning embeds graph data into curved manifolds to better preserve non-Euclidean structures, especially hierarchical, cyclic, or densely connected patterns~\cite{nickel2017poincare,chami2019hyperbolic,wilson2014spherical,liu2019hyperbolic}. Hyperbolic and curvature-aware graph models have shown the benefit of geometric inductive biases for graph representation learning~\cite{zhang2021hyperbolic,ace-hgnn,curvgan,guo2025graphmore}. Yet most existing methods focus on a fixed graph domain or a limited geometric assumption. In contrast, we combine Euclidean, hyperbolic, and spherical priors through a mixture of Riemannian experts, allowing the model to adapt its geometry to unseen graph anomaly patterns.
\section{Problem Formulation}
\label{sec:problem}

We consider node-level graph anomaly detection on an attributed graph $\mathcal{G}=(\mathcal{V},\mathcal{E},\mathbf{X}_0)$ with adjacency matrix $\mathbf{A}$. Let $\mathcal{V}_a$ and $\mathcal{V}_n$ denote anomalous and normal nodes, where $|\mathcal{V}_a| \ll |\mathcal{V}_n|$. The goal is to learn an anomaly scoring function $f:\mathcal{V}\rightarrow\mathbb{R}$ that assigns higher scores to anomalous nodes.

In the generalizable setting, a model is trained on source graphs $\mathcal{D}_{\mathrm{train}}=\{\mathcal{G}^{(1)},\ldots,\mathcal{G}^{(M)}\}$ and evaluated on disjoint unseen target graphs $\mathcal{D}_{\mathrm{test}}=\{\mathcal{G}^{(M+1)},\ldots,\mathcal{G}^{(P)}\}$. We focus on zero-shot transfer, where target-graph labels and fine-tuning are unavailable. Since raw feature dimensions vary across graphs, each $\mathbf{X}_0^{(i)}$ is projected into a shared representation $\mathbf{X}^{(i)}\in\mathbb{R}^{N_i\times D}$.
\section{The Proposed \modelname~Framework}
\label{sec:method}

\begin{figure*}[ht!]
    \centering
    \includegraphics[width=\textwidth]{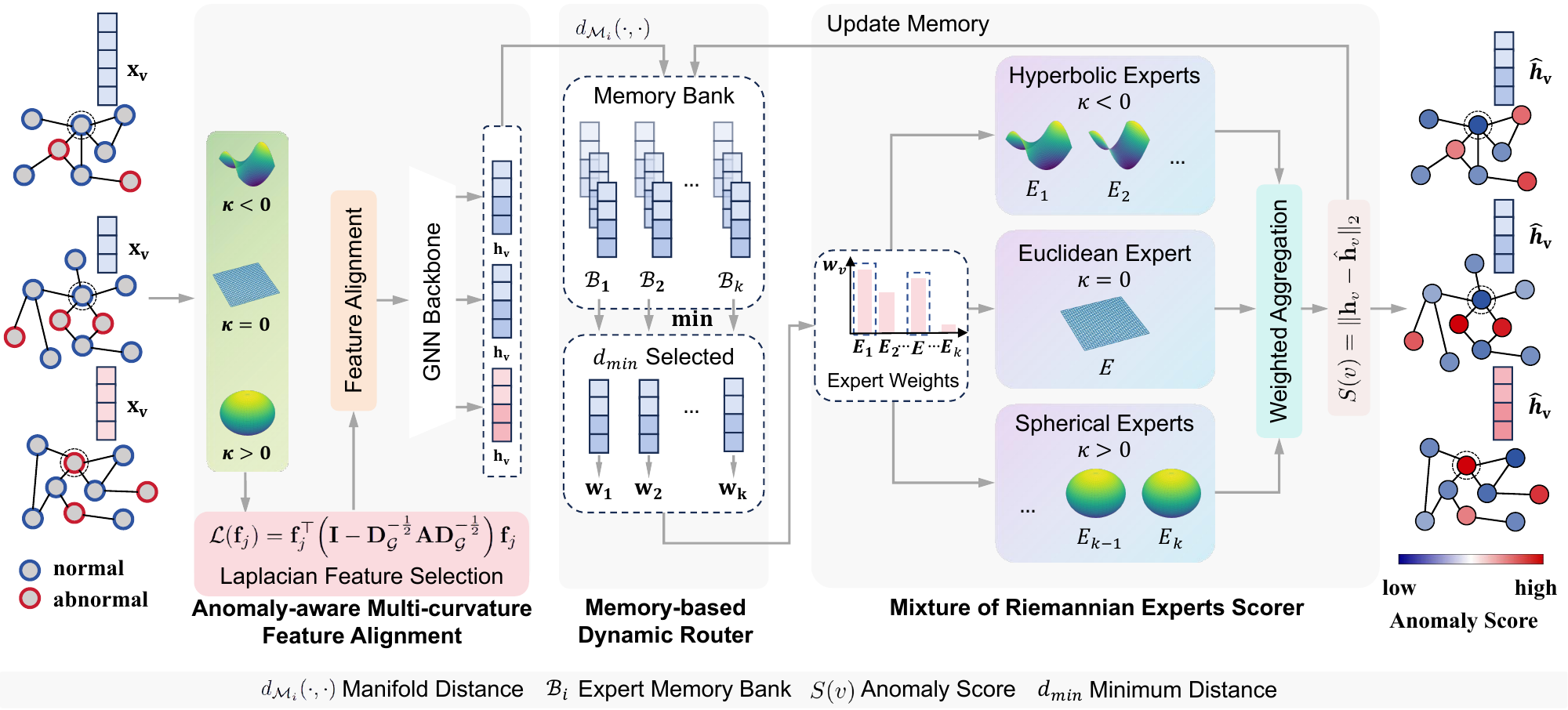} 
    \caption{The overall architecture of the \modelname~framework. 
    (1) The Anomaly-aware Multi-curvature Feature Alignment module projects raw features into a unified, geometry-aware representation. 
    (2) The Mixture of Riemannian Experts Scorer utilizes a GNN backbone to encode structural context and employs specialized Riemannian expert networks. 
    (3) The Memory-based Dynamic Router adaptively assigns nodes to these experts based on historical reconstruction quality to compute the final anomaly score.}
    \label{fig:framework}
\end{figure*}

Graph anomaly patterns exhibit geometric heterogeneity across domains, making a single fixed-curvature space insufficient for zero-shot generalizable GAD. To address this issue, we propose \modelname, a geometric-aware framework with a mixture-of-experts architecture. As illustrated in Figure~\ref{fig:framework}, \modelname{} consists of three coordinated components:
\begin{itemize}[leftmargin=*]
\item \textit{Anomaly-aware Multi-curvature Feature Alignment:} This module projects raw node features into parallel curvature spaces and constructs a unified geometry-aware representation.
\item \textit{Mixture of Riemannian Experts Scorer:} This module encodes structural context and reconstructs node embeddings through specialized experts operating in distinct curvature spaces.
\item \textit{Memory-based Dynamic Router:} This module assigns nodes to suitable experts according to historical reconstruction quality, improving expert selection under cross-graph distribution shifts.
\end{itemize}
Together, these components allow \modelname{} to adapt its geometric representation to diverse anomaly patterns in unseen graphs.

\subsection{Anomaly-aware Multi-curvature Feature Alignment}

Raw node features extracted from real-world graphs are inherently geometry-agnostic, as they exist in a flat Euclidean space without explicit geometric information. 
Yet, to effectively capture diverse structural patterns without distortion, our Riemannian expert networks must operate in distinct constant-curvature spaces. 

This creates a fundamental mismatch: the Riemannian experts in MoE require manifold-aware inputs, while raw features lack geometric semantics. 
Simply feeding Euclidean features directly to curved-space experts is geometrically inconsistent and fails to leverage the representational advantages of non-Euclidean geometries. 
Moreover, different graph domains exhibit preferences for different geometric curvatures, as demonstrated empirically in Section~\ref{sec:intro}. 

To construct robust and generalizable inputs for Riemannian experts, we explicitly embed signals from multiple Riemannian spaces into the initial node features, preserving both the rich attribute information and global structural consistency.

To address this challenge, we introduce a feature alignment module that explicitly projects raw features into diverse Riemannian spaces before feeding them to the experts. 
Given a raw feature matrix $\mathbf{X}_0$, the module performs a parallel transformation into a set of diverse Riemannian spaces, $\{\mathcal{M}_{\kappa_c}\}_{c=1}^C$. 
This set includes Euclidean space ($\kappa=0$) alongside multiple curved spaces characterized by different constant negative ($\kappa<0$) and positive curvatures ($\kappa>0$). 
The total target feature dimension $D$ is partitioned across these $C$ spaces, such that $D = \sum_{c=1}^C D_c$, where $D_c$ is the final dimensionality assigned to the $c$-th space. 
For curved spaces, the raw features are first mapped into the tangent space at the origin before further processing.
To ensure that the features selected for each Riemannian space are both informative and dimensionally consistent, we employ a two-stage selection process:

\textbf{Stage 1: Dimensionality Reduction. }
To construct geometry-specific feature candidates for each target manifold $\mathcal{M}_{c}$, we first project the raw features $\mathbf{X}_0$ into $C$ parallel constant-curvature manifolds $\{\mathcal{M}_{\kappa_c}\}_{c=1}^C$. 
Specifically, for curved spaces, the Euclidean attributes are represented in the tangent space at the origin before dimensionality reduction. We then apply Principal Component Analysis (PCA) to obtain compact feature candidates for each parallel branch.
Then, we perform Principal Component Analysis (PCA)~\cite{pearson1901liii,hotelling1933analysis} on these tangent-space features to extract the principal components corresponding to each specific $c$-th curvature space. 
This step identifies the principal components of the feature space, capturing global high-variance attribute signals that are most informative for node discrimination.

\textbf{Stage 2: Laplacian Feature Selection.} 
We then compute the Laplacian score for each feature $\mathbf{f}_j$ in the intermediate set $\mathbf{X}_c'$:
\begin{equation}
    \mathcal{L}(\mathbf{f}_j) = \mathbf{f}_j^\top \!\left(\mathbf{I} - \mathbf{D}_{\mathcal{G}}^{-\frac{1}{2}} \mathbf{A} \mathbf{D}_{\mathcal{G}}^{-\frac{1}{2}}\right) \mathbf{f}_j,
    \label{eq:laplacian_score}
\end{equation}
where $\mathbf{D}_{\mathcal{G}}=\operatorname{diag}(\mathbf{A}\mathbf{1})$. 
Note that we omit the denominator term $\mathbf{f}_j^\top \mathbf{D}_{\mathcal{G}} \mathbf{f}_j$ commonly used in the classical definition, consistent with our implementation. 
We then select the top-$D_c$ features with the lowest scores to form the final feature set for that space, $\mathbf{X}_c^{\mathrm{final}} \in \mathbb{R}^{N \times D_c}$. 
This selection ensures that the chosen features are smooth and consistent with the local graph topology, thus preserving crucial structural information essential for anomaly detection.

Ideally, PCA extracts high-variance attribute signals, while the subsequent Laplacian Score selection enforces structural consistency, yielding features that are both informative and topology-consistent. 
Finally, the selected features from all Riemannian spaces are concatenated to form a unified geometry-aware feature matrix $\mathbf{X} = \mathrm{CONCAT}(\mathbf{X}_1^{\mathrm{final}}, \dots, \mathbf{X}_C^{\mathrm{final}}) \in \mathbb{R}^{N \times D}$.


\begin{algorithm}[!t]
\caption{The overall training pipeline of \modelname.}
\label{alg:training}
\begin{algorithmic}[1]
\REQUIRE Source graphs $\mathcal{D}_{\text{train}}$, number of experts $K$, active experts $k$, GNN hops $L$, loss weights $\{\lambda_j\}_{j=1}^5$, cold-start epochs $E_{\text{cold}}$.
\ENSURE Trained model parameters $\Theta$ and expert memories $\{\mathcal{B}_j\}_{j=1}^K$.
\FOR{$i = 1,\dots,M$}
    \STATE Align features: $\mathbf{X}^{(i)} \leftarrow \text{MCFA}(\mathbf{X}_0^{(i)}, \mathbf{A}^{(i)})$.
    \STATE Propagate structure: $\{\mathbf{H}^{(l)}\}_{l=0}^L \leftarrow \text{Prop}(\mathbf{X}^{(i)}, \mathbf{A}^{(i)}, L)$.
\ENDFOR
\STATE Initialize expert memories $\mathcal{B}_j \leftarrow \emptyset$ for $j = 1, \ldots, K$.
\FOR{$\text{epoch} = 1,\dots,E_{\text{total}}$}
    \FOR{each $\mathcal{G}^{(i)} \in \mathcal{D}_{\text{train}}$}
        \STATE Encode structural context: $\mathbf{H}_{res} \leftarrow \text{GNN}(\{\mathbf{H}^{(l)}\}_{l=0}^L)$.
        \STATE Compute routing logits using adaptive exploration and the memory-based score in Eq.~\eqref{eq:routing_score}.
        \STATE Select top-k experts $\mathcal{S}(v)$ and weights $\mathbf{w}_v$ by Eq.~\eqref{eq:gating_softmax}.
        \STATE Reconstruct embeddings $\hat{\mathbf{H}}$ by Eq.~\eqref{eq:moe_output} and decode $\hat{\mathbf{X}},\hat{\mathbf{A}}$.
        \STATE Compute $\mathcal{L}_{\text{embed}}$, $\mathcal{L}_{\text{feat}}$, $\mathcal{L}_{\text{struct}}$, $\mathcal{L}_{\text{con}}$, and $\mathcal{L}_{\text{gate}}$.
        \STATE $\mathcal{L}_{\text{total}} \leftarrow \sum_{j=1}^5 \lambda_j \mathcal{L}_j$ by Eq.~\eqref{eq:loss_total}.
        \STATE Update parameters: $\Theta \leftarrow \Theta - \eta \nabla_{\Theta} \mathcal{L}_{\text{total}}$.
        \IF{$\text{epoch} > E_{\text{cold}}$}
            \FOR{$j=1,\dots,K$}
                \STATE Update $\mathcal{B}_j$ using $q(v,j)$ and the memory-admission rule.
            \ENDFOR
        \ENDIF
    \ENDFOR
\ENDFOR
\end{algorithmic}
\end{algorithm}

\subsection{Mixture of Riemannian Experts Scorer}

Different graph domains exhibit fundamentally different structural characteristics that are best captured in distinct Riemannian spaces. 
For instance, hierarchical graphs with tree-like structures are naturally suited for hyperbolic geometry, while graphs with dense, cyclical communities align better with positively curved spherical spaces. 
When anomalies are defined as deviations from these underlying structural patterns, enforcing all graphs into a single fixed-curvature space induces critical distortions. 
In a hyperbolic space, dense communities may explode and distort, making anomalous isolated nodes indistinguishable from the distorted normal nodes. 
Conversely, in a spherical space, hierarchical structures may be compressed, obscuring bridge nodes that improperly connect distant branches. 
This geometric mismatch makes it extremely difficult to distinguish subtle structural anomalies from normal variations. 
Effectively capturing diverse structural patterns without distortion requires an adaptive mechanism that represents nodes in their most suitable Riemannian space. 
This necessitates a unified architecture capable of accommodating multiple, potentially conflicting Riemannian representations. 
In the zero-shot setting, anomaly patterns across graph domains prefer different Riemannian representations, so a single learned curvature becomes a compromise, whereas a MoE formulation enables adaptive geometric selection. 

Node embeddings, which encode both multi-curvature attributes and structural context from the GNN backbone, are fed into the central innovation of \modelname: the mixture of Riemannian experts scorer. 
The GNN backbone first produces multi-hop feature matrices $\{\mathbf{H}^{(0)}, \dots, \mathbf{H}^{(K_{\mathrm{hops}})}\}$ where $\mathbf{H}^{(0)}=\mathbf{X}$. 
A shared MLP $\Phi$ transforms $\{\mathbf{H}^{(0)}, \dots, \mathbf{H}^{(K_{\mathrm{hops}})}\}$ into $\{\mathbf{X}'_k=\Phi(\mathbf{H}^{(k)})\}$. 
The final node embedding $\mathbf{h}_v$ is then constructed using a residual formulation to capture structural deviations, defined as $\mathbf{h}_v = \mathrm{CONCAT}(\mathbf{X}'_1(v) - \mathbf{X}'_0(v), \dots, \mathbf{X}'_{K_{\mathrm{hops}}}(v) - \mathbf{X}'_0(v))$.
This residual formulation is designed to explicitly capture how a node's representation evolves as structural information is aggregated from progressively larger neighborhoods. 
Large deviations between a node's initial features ($\mathbf{X}'_0$) and its aggregated features ($\mathbf{X}'_k$) indicate structural irregularity, providing a strong signal for anomaly detection. 
This embedding $\mathbf{h}_v$ is then input into the MoE scorer.
In this way, the scorer receives node representations that reflect both attribute-level multi-curvature cues and topology-induced structural deviations.

To simultaneously accommodate multiple Riemannian representations, we employ a Mixture-of-Experts (MoE) architecture with specialized Riemannian Expert Networks. 
The MoE model consists of a set of $K$ expert networks, $\{E_i\}_{i=1}^K$. 
Each expert is a Riemannian expert, a small neural network designed to operate within a specific Riemannian manifold $\mathcal{M}_{\kappa_i}$ with learnable curvature $\kappa_i$. 
Each expert $E_i$ takes a node embedding $\mathbf{h}_v$ from the Euclidean tangent space and computes its reconstruction $\hat{\mathbf{h}}_v^{(i)}$. 
This process involves mapping the input onto the manifold, processing it through manifold-aware layers, and mapping the result back to the tangent space:
\begin{equation}
\hat{\mathbf{h}}_v^{(i)} = \log_{o_i}\left(E_i\left(\exp_{o_i}\left(\mathbf{h}_v\right)\right)\right),
\label{eq:expert_recon}
\end{equation}
where $\exp_{o_i}$ and $\log_{o_i}$ are the exponential and logarithmic maps at the origin $o_i$ of the manifold $\mathcal{M}_{\kappa_i}$. 
This design allows different experts to specialize in reconstructing nodes that conform to different geometric priors, effectively capturing the diverse structural patterns present across graph domains.
Following the practice of modern MoE, we employ sparse top-$k$ routing to enable efficient conditional computation. 
For each node, the router first computes the gating weights $\mathbf{g}(\mathbf{h}_v)$ from the raw logits $\mathbf{s}(\mathbf{h}_v)$:
\begin{equation}
\mathbf{g}(\mathbf{h}_v) = \operatorname{Softmax}\left(\frac{\mathbf{s}(\mathbf{h}_v)}{\tau}\right),
\label{eq:gating_softmax}
\end{equation}
where the temperature is controlled by the parameter $\tau$. From the computed probabilities, the top-$k$ experts with maximal influence are selected for the active set $\mathcal{S}(v)$. The final output $\hat{\mathbf{h}}_v$ is computed as a weighted linear combination of outputs from these chosen experts, re-normalized among the $k$ candidates:
\begin{equation}
\hat{\mathbf{h}}_v = \sum_{i \in \mathcal{S}(v)} \frac{\exp\left(s_i(\mathbf{h}_v)/\tau\right)}{\sum_{j \in \mathcal{S}(v)} \exp\left(s_j(\mathbf{h}_v)/\tau\right)} \cdot \hat{\mathbf{h}}_v^{(i)}.
\label{eq:moe_output}
\end{equation}


\subsection{Memory-based Dynamic Router}
With specialized Riemannian Expert Networks, each tailored to distinct structural patterns under its optimal curvature, the central routing challenge is to assign each node to the most suitable Riemannian expert for accurate anomaly scoring.
Conventional routing mechanisms in MoE models typically rely on learned gating networks that predict expert assignments based on input features alone. 
For anomaly detection, however, the routing policy must be tightly coupled to the reconstruction objective: nodes should be directed to experts who can reconstruct their patterns with high fidelity, as reconstruction error directly determines the anomaly score. 
Without this alignment, the router may assign nodes to geometrically inappropriate experts, leading to unreliable anomaly scores and degraded detection performance. 
Moreover, a node's optimal Riemannian space depends on its historical reconstruction quality across different experts, which is not captured by feature-based routing alone. 
This motivates a routing mechanism capable of adaptively selecting experts based on their demonstrated ability to reconstruct similar patterns, thereby ensuring that the routing policy directly aligns with the anomaly detection objective.

To enable reconstruction quality-based routing, we propose a memory-based dynamic router that makes routing decisions based on historical performance patterns. 
Each expert $E_i$ is equipped with an expert memory bank $\mathcal{B}_i$, which stores a collection of node embeddings $\{\mathbf{m}_{i,1}, \mathbf{m}_{i,2}, \dots\}$ that the expert has previously reconstructed with high fidelity. 
These memory banks serve as geometric prototypes, representing the patterns that each expert specializes in handling.
For a given node embedding $\mathbf{h}_v$, the router computes its similarity to each memory bank. 
The distance-based score $r_i(\mathbf{h}_v)$ for expert $i$ is defined as the negative minimum manifold distance from $\mathbf{h}_v$ to any embedding in the memory bank $\mathcal{B}_i$:
\begin{equation}
r_i(\mathbf{h}_v) = - \min_{\mathbf{m} \in \mathcal{B}_i} d_{\mathcal{M}_i}\left(\exp_{o_i}\left(\mathbf{h}_v\right), \exp_{o_i}\left(\mathbf{m}\right)\right),
\label{eq:routing_score}
\end{equation}
where $d_{\mathcal{M}_i}(\cdot, \cdot)$ is the distance metric on the manifold of expert $i$. 
The resulting score vector $\mathbf{r}$ is linearly projected to obtain the routing logits $\mathbf{s}$.
This score measures how close node $v$ is to the patterns that expert $i$ is specialized in handling based on its historical reconstruction performance.

Expert Memory Banks are not static but are populated and refined dynamically throughout the training process to store a representative set of node embeddings that each expert reconstructs with high fidelity. 
The update mechanism is governed by a multi-stage strategy involving a cold-start period, a dynamic quality gate, and a strategic replacement policy.
The update process is triggered after each training batch. 
To allow the expert networks to stabilize before populating memory banks, we implement a cold-start period during which memory banks remain inactive. 
During training, the router probabilistically switches between near-uniform exploration logits and memory-based logits using an adaptive exploration probability.
This prevents the storage of low-quality reconstructions from poorly initialized experts.
After the cold-start period, for each expert $E_i$, we calculate a quality score $q(v, i)$ for each candidate node embedding $\mathbf{h}_v$ assigned to it. 

Let $\mathcal{L}_{\mathrm{recon}}(v,i)
= \|\mathbf{h}_v-\hat{\mathbf{h}}_v^{(i)}\|_2^2$
denote the reconstruction error. For the nodes assigned to expert $E_i$
in the current batch $\mathcal{B}_{\mathrm{batch}}^{(i)}$, we define
\begin{equation}
q(v,i)=1-
\frac{\mathcal{L}_{\mathrm{recon}}(v,i)}
{\max_{v'\in\mathcal{B}_{\mathrm{batch}}^{(i)}}
\mathcal{L}_{\mathrm{recon}}(v',i)+\epsilon}.
\label{eq:quality_calc}
\end{equation}
We retain candidates with $q(v,i)\geq0.7$, falling back to the
70th-percentile threshold if none qualify. After the cold-start period,
the memory-admission threshold increases from $0.3$ to $0.7$ over
10 epochs. At most three candidates are added per expert per update;
when a bank is full, a candidate replaces its lowest-quality entry
only if its quality improves by more than $0.1$.

By adaptively assigning nodes based on historical reconstruction quality rather than feature similarity alone, the routing policy aligns expert selection with the anomaly detection objective, directing each node to the Riemannian space where it is best modeled.

\subsection{Model Training and Optimization}
\modelname~is trained end-to-end in an unsupervised manner, as detailed in Algorithm~\ref{alg:training}. The total loss function $\mathcal{L}_{\mathrm{total}}$ is a weighted sum of multiple components:
\begin{itemize}
    \item \textit{Embedding Reconstruction Loss ($\mathcal{L}_{\mathrm{embed}}$):} The primary reconstruction objective, defined as the Mean Squared Error:
\begin{equation}
    \mathcal{L}_{\mathrm{embed}} = \frac{1}{|\mathcal{V}|} \sum_{v \in \mathcal{V}} \|\mathbf{h}_v - \hat{\mathbf{h}}_v\|_2^2.
    \label{eq:loss_embed}
\end{equation}
 \item \textit{Feature Reconstruction Loss ($\mathcal{L}_{\mathrm{feat}}$):} This loss enforces the reconstructed embedding to preserve attribute information. 
A linear decoder maps the reconstructed embedding $\hat{\mathbf{h}}_v$ back to the original feature space, and the Mean Squared Error is computed against the original features $\mathbf{X}_v$:
\begin{equation}
    \mathcal{L}_{\mathrm{feat}} = \frac{1}{|\mathcal{V}|} \sum_{v \in \mathcal{V}} \|\mathbf{X}_v - \mathrm{Decoder}_{\mathrm{feat}}\left(\hat{\mathbf{h}}_v\right)\|_2^2.
    \label{eq:loss_feat}
\end{equation}
\item \textit{Structure Reconstruction Loss ($\mathcal{L}_{\mathrm{struct}}$):} 
This loss preserves topological information. 
The adjacency matrix is reconstructed via an inner product of the reconstructed embeddings. 
Let $\hat{\mathbf{H}}$ be the matrix formed by stacking all reconstructed node embeddings $\hat{\mathbf{h}}_v$. 
The Binary Cross-Entropy (BCE) loss is then computed against the original adjacency matrix $\mathbf{A}$:
\begin{equation}
    \mathcal{L}_{\mathrm{struct}} = \operatorname{BCE}\left(\sigma\left(\hat{\mathbf{H}}\hat{\mathbf{H}}^\top\right), \mathbf{A}\right).
    \label{eq:loss_struct}
\end{equation}
\item \textit{Structure-Contrastive Loss ($\mathcal{L}_{\mathrm{con}}$):} 
This structure-aware InfoNCE loss encourages connected nodes to have similar representations. 
For a node $v$, the loss is:
\begin{equation}
    \mathcal{L}_{\mathrm{con}}(v) = -\log \frac{\sum_{u \in \mathcal{P}_v} \exp\left(\mathrm{sim}\left(\mathbf{h}_v, \mathbf{h}_u\right)/\tau_c\right)}{\sum_{j \in \mathcal{V}} \exp\left(\mathrm{sim}\left(\mathbf{h}_v, \mathbf{h}_j\right)/\tau_c\right)},
    \label{eq:loss_con}
\end{equation}
where $\mathcal{P}_v = \{u \mid \mathbf{A}_{vu}=1\}$ is the set of positive neighbors of $v$, and $\tau_c$ is a temperature parameter. 
For large graphs, the denominator is computed over a randomly sampled subset of nodes, consistent with our implementation.
\item \textit{Gate Entropy Regularization ($\mathcal{L}_{\mathrm{gate}}$):} 
This entropy-based loss penalizes uncertain routing and encourages more confident expert assignments. We define the gate loss as the average routing entropy:
\begin{align}
    \mathcal{L}_{\mathrm{gate}}
    &= \frac{1}{|\mathcal{V}|} \sum_{v \in \mathcal{V}}
    H\left(\mathbf{g}\left(\mathbf{h}_v\right)\right) \nonumber \\
    &= -\frac{1}{|\mathcal{V}|} \sum_{v \in \mathcal{V}} \sum_{i=1}^{K}
    g_i\left(\mathbf{h}_v\right)
    \log g_i\left(\mathbf{h}_v\right).
    \label{eq:loss_gate}
\end{align}
\end{itemize}

By integrating these modules with specific hyperparameters, the total training loss is established as follows:
\begin{equation} \label{eq:loss_total}
\begin{split}
    \mathcal{L}_{\mathrm{total}} ={}& \lambda_1 \mathcal{L}_{\mathrm{embed}} + \lambda_2 \mathcal{L}_{\mathrm{feat}} + \lambda_3 \mathcal{L}_{\mathrm{struct}} \\
    & + \lambda_4 \mathcal{L}_{\mathrm{con}} + \lambda_5 \mathcal{L}_{\mathrm{gate}},
\end{split}
\end{equation}
where $\lambda_1, \lambda_2, \lambda_3, \lambda_4, \lambda_5$ are the hyperparameters that control the relative contributions of each loss term.

For the structure-contrastive objective, we use a memory-efficient implementation with mini-batch sampling and chunked computations to approximate InfoNCE on large graphs without exceeding GPU memory. 
Specifically, the similarity function $\mathrm{sim}(\mathbf{h}_v, \mathbf{h}_u)$ operates on the node embedding vectors.

\subsection{Anomaly Inference}
At inference time, \modelname~is deployed on unseen graphs in a zero-shot manner. For each node $v$, we compute the anomaly score $S(v)$ as the $L_2$ norm of the residual between the input representation $\mathbf{h}_v$ and its MoE-reconstructed counterpart $\hat{\mathbf{h}}_v$:
\begin{equation}
S(v) = \|\mathbf{h}_v - \hat{\mathbf{h}}_v\|_2.
\label{eq:anomaly_score}
\end{equation}
Higher scores indicate greater deviation from curvature-aware normal patterns learned by the Riemannian experts, reflecting reduced reconstruction fidelity and thus stronger evidence of anomalous structure.

\begin{table}[t]
    \centering
    \caption{Statistics of the benchmark datasets.}
    \label{tab:datasets}
    \resizebox{0.95\linewidth}{!}{\begin{tabular}{lrrrr}
        \toprule
        \textbf{Dataset} & \textbf{\#Nodes} & \textbf{\#Edges} & \textbf{\#Feat.} & \textbf{\#Anom. (\%)} \\
        \midrule
        \multicolumn{5}{l}{\textit{Training Datasets}} \\
        PubMed & 19,717 & 44,338 & 500 & 600 (3.04) \\
        Flickr & 7,575 & 239,738 & 12,047 & 450 (5.94) \\
        Reddit & 10,984 & 168,016 & 64 & 366 (3.33) \\
        YelpChi & 23,831 & 49,315 & 32 & 1,217 (5.10) \\
        \midrule
        \multicolumn{5}{l}{\textit{Test Datasets}} \\
        ACM & 16,484 & 71,980 & 8,337 & 597 (3.62) \\
        Amazon & 10,244 & 175,608 & 25 & 693 (6.76) \\
        BlogCatalog & 5,196 & 171,743 & 8,189 & 300 (5.77) \\
        Citeseer & 3,327 & 4,732 & 3,703 & 150 (4.50) \\
        Cora & 2,708 & 5,429 & 1,433 & 150 (5.53) \\
        Facebook & 1,081 & 55,104 & 576 & 25 (2.31) \\
        Weibo & 8,405 & 407,963 & 400 & 868 (10.30) \\
        \bottomrule
    \end{tabular}
    }
\end{table}

\begin{table*}[htbp]
  \centering
  \caption{AUROC results on target GAD datasets. ARC is reported under a 10-shot target-label setting; all other methods are evaluated without target labels. Best and second-best results are bolded and underlined.}
  \tabcolsep=0.15cm
  \resizebox{\linewidth}{!}{%
    \begin{tabular}{c|c|r|r|r|r|r|r|r|r}
    \toprule
    \multicolumn{2}{c|}{Dataset} & \multicolumn{3}{c|}{\textit{Citation}} & \multicolumn{3}{c|}{\textit{Social}} & \multicolumn{1}{c|}{\textit{Co-review}} & \multicolumn{1}{c}{\multirow{2}[4]{*}{Average}} \\
    \cmidrule(r){1-2} \cmidrule(lr){3-5} \cmidrule(lr){6-8} \cmidrule(l){9-9} 
    \multicolumn{2}{c|}{Method} & \multicolumn{1}{c|}{ACM} & \multicolumn{1}{c|}{Citeseer} & \multicolumn{1}{c|}{Cora} & \multicolumn{1}{c|}{BlogCatalog} & \multicolumn{1}{c|}{Facebook} & \multicolumn{1}{c|}{Weibo} & \multicolumn{1}{c|}{Amazon} &  \\
    \midrule
    \multirow{4}[2]{*}{\textit{Supervised}} & GCN   & 47.69\ensuremath{\pm}\scalebox{0.7}{1.48} & 53.83\ensuremath{\pm}\scalebox{0.7}{4.01} & 51.99\ensuremath{\pm}\scalebox{0.7}{3.20} & 43.83\ensuremath{\pm}\scalebox{0.7}{2.39\phantom{0}} & 46.09\ensuremath{\pm}\scalebox{0.7}{18.88} & 37.64\ensuremath{\pm}\scalebox{0.7}{13.84} & 41.10\ensuremath{\pm}\scalebox{0.7}{4.24} & 46.02\phantom{0} \\
          & GAT   & 46.54\ensuremath{\pm}\scalebox{0.7}{4.71} & 44.58\ensuremath{\pm}\scalebox{0.7}{8.25} & 44.65\ensuremath{\pm}\scalebox{0.7}{6.20} & 49.32\ensuremath{\pm}\scalebox{0.7}{1.70\phantom{0}} & 53.96\ensuremath{\pm}\scalebox{0.7}{26.14} & 35.50\ensuremath{\pm}\scalebox{0.7}{6.91\phantom{0}} & 55.35\ensuremath{\pm}\scalebox{0.7}{5.69} & 47.13\phantom{0} \\
          & BWGNN & 67.43\ensuremath{\pm}\scalebox{0.7}{2.48} & 42.58\ensuremath{\pm}\scalebox{0.7}{2.86} & 39.75\ensuremath{\pm}\scalebox{0.7}{3.16} & 69.91\ensuremath{\pm}\scalebox{0.7}{4.66\phantom{0}} & 64.84\ensuremath{\pm}\scalebox{0.7}{4.17\phantom{0}} & 50.08\ensuremath{\pm}\scalebox{0.7}{0.03\phantom{0}} & 50.01\ensuremath{\pm}\scalebox{0.7}{0.04} & 54.94\phantom{0} \\
          & GHRN  & 52.72\ensuremath{\pm}\scalebox{0.7}{1.01} & 54.11\ensuremath{\pm}\scalebox{0.7}{0.88} & 47.65\ensuremath{\pm}\scalebox{0.7}{2.03} & 62.28\ensuremath{\pm}\scalebox{0.7}{2.42\phantom{0}} & 54.04\ensuremath{\pm}\scalebox{0.7}{3.81\phantom{0}} & 17.79\ensuremath{\pm}\scalebox{0.7}{5.00\phantom{0}} & 63.52\ensuremath{\pm}\scalebox{0.7}{1.92} & 50.30\phantom{0} \\
    \midrule
    \multirow{3}[2]{*}{\textit{Unsupervised}} & AnomalyDAE & 77.60\ensuremath{\pm}\scalebox{0.7}{1.03} & 82.00\ensuremath{\pm}\scalebox{0.7}{2.78} & 84.01\ensuremath{\pm}\scalebox{0.7}{1.19} & 57.53\ensuremath{\pm}\scalebox{0.7}{0.30\phantom{0}} & 10.87\ensuremath{\pm}\scalebox{0.7}{0.70\phantom{0}} & 19.77\ensuremath{\pm}\scalebox{0.7}{1.14\phantom{0}} & 58.38\ensuremath{\pm}\scalebox{0.7}{0.25} & 55.74\phantom{0} \\
          & CoLA  & 49.30\ensuremath{\pm}\scalebox{0.7}{1.00} & 52.56\ensuremath{\pm}\scalebox{0.7}{1.36} & 49.35\ensuremath{\pm}\scalebox{0.7}{1.52} & 49.50\ensuremath{\pm}\scalebox{0.7}{1.27\phantom{0}} & 50.70\ensuremath{\pm}\scalebox{0.7}{5.49\phantom{0}} & 45.25\ensuremath{\pm}\scalebox{0.7}{2.39\phantom{0}} & 47.57\ensuremath{\pm}\scalebox{0.7}{3.02} & 49.17\phantom{0} \\
          & TAM   & 50.39\ensuremath{\pm}\scalebox{0.7}{0.08} & 46.44\ensuremath{\pm}\scalebox{0.7}{0.15} & 48.78\ensuremath{\pm}\scalebox{0.7}{0.17} & 33.03\ensuremath{\pm}\scalebox{0.7}{0.28\phantom{0}} & 54.56\ensuremath{\pm}\scalebox{0.7}{0.36\phantom{0}} & 21.50\ensuremath{\pm}\scalebox{0.7}{0.30\phantom{0}} & 52.37\ensuremath{\pm}\scalebox{0.7}{0.12} & 43.87\phantom{0} \\
    \midrule
    \multirow{5}[2]{*}{\textit{Generalist}} & UNPrompt & 73.20\ensuremath{\pm}\scalebox{0.7}{0.36} & 59.53\ensuremath{\pm}\scalebox{0.7}{0.53} & 63.18\ensuremath{\pm}\scalebox{0.7}{1.29} & 70.85\ensuremath{\pm}\scalebox{0.7}{0.50\phantom{0}} & 60.64\ensuremath{\pm}\scalebox{0.7}{1.94\phantom{0}} & 66.29\ensuremath{\pm}\scalebox{0.7}{1.38\phantom{0}} & \underline{72.14}\ensuremath{\pm}\scalebox{0.7}{2.94} & 66.55\phantom{0} \\
          & ARC (10-shot) & 78.33\ensuremath{\pm}\scalebox{0.7}{1.13} & 82.31\ensuremath{\pm}\scalebox{0.7}{2.89} & 82.22\ensuremath{\pm}\scalebox{0.7}{2.34} & 72.59\ensuremath{\pm}\scalebox{0.7}{1.10\phantom{0}} & 64.94\ensuremath{\pm}\scalebox{0.7}{3.72} & \underline{88.65}\ensuremath{\pm}\scalebox{0.7}{0.82} & 57.95\ensuremath{\pm}\scalebox{0.7}{6.13} & 75.28\phantom{0} \\
          & AnomalyGFM & 38.91\ensuremath{\pm}\scalebox{0.7}{1.21} & 40.37\ensuremath{\pm}\scalebox{0.7}{1.37} & 40.93\ensuremath{\pm}\scalebox{0.7}{1.09} & 40.93\ensuremath{\pm}\scalebox{0.7}{1.22\phantom{0}} & \underline{74.45}\ensuremath{\pm}\scalebox{0.7}{2.36\phantom{0}} & 44.43\ensuremath{\pm}\scalebox{0.7}{4.89\phantom{0}} & 61.24\ensuremath{\pm}\scalebox{0.7}{5.21} & 48.75\phantom{0} \\
          & IA-GGAD & \underline{78.44}\ensuremath{\pm}\scalebox{0.7}{0.46} & \textbf{91.54}\ensuremath{\pm}\scalebox{0.7}{0.32} & \underline{85.69}\ensuremath{\pm}\scalebox{0.7}{0.77} & \underline{73.03}\ensuremath{\pm}\scalebox{0.7}{1.20\phantom{0}} & 69.59\ensuremath{\pm}\scalebox{0.7}{1.39\phantom{0}} & 87.35\ensuremath{\pm}\scalebox{0.7}{0.40\phantom{0}} & 53.38\ensuremath{\pm}\scalebox{0.7}{1.76} & \underline{77.00}\phantom{0} \\
          & \textbf{\modelname} & \textbf{81.17}\ensuremath{\pm}\scalebox{0.7}{0.05} & \underline{90.28}\ensuremath{\pm}\scalebox{0.7}{0.13} & \textbf{86.39}\ensuremath{\pm}\scalebox{0.7}{0.20} & \textbf{73.09}\ensuremath{\pm}\scalebox{0.7}{0.06\phantom{0}} & \textbf{75.75}\ensuremath{\pm}\scalebox{0.7}{0.95\phantom{0}} & \textbf{91.03}\ensuremath{\pm}\scalebox{0.7}{0.11\phantom{0}} & \textbf{76.90}\ensuremath{\pm}\scalebox{0.7}{1.80} & \textbf{82.09}\phantom{0} \\
    \bottomrule
    \end{tabular}%
  }
  \label{tab:mainauc}%
\end{table*}
\begin{table*}[htbp]
  \centering
  \caption{AUPRC results on target GAD datasets. ARC is reported under a 10-shot target-label setting; all other methods are evaluated without target labels. Best and second-best results are bolded and underlined.}
  \tabcolsep=0.15cm
  \resizebox{\linewidth}{!}{%
    \begin{tabular}{c|c|r|r|r|r|r|r|r|r}
    \toprule
    \multicolumn{2}{c|}{Dataset} & \multicolumn{3}{c|}{\textit{Citation}} & \multicolumn{3}{c|}{\textit{Social}} & \multicolumn{1}{c|}{\textit{Co-review}} & \multicolumn{1}{c}{\multirow{2}[4]{*}{Average}} \\
    \cmidrule(r){1-2} \cmidrule(lr){3-5} \cmidrule(lr){6-8} \cmidrule(l){9-9} 
    \multicolumn{2}{c|}{Method} & \multicolumn{1}{c|}{ACM} & \multicolumn{1}{c|}{Citeseer} & \multicolumn{1}{c|}{Cora} & \multicolumn{1}{c|}{BlogCatalog} & \multicolumn{1}{c|}{Facebook} & \multicolumn{1}{c|}{Weibo} & \multicolumn{1}{c|}{Amazon} &  \\
    \midrule
    \multirow{4}[2]{*}{\textit{Supervised}} & GCN   & 3.68\ensuremath{\pm}\scalebox{0.7}{0.16} & 5.21\ensuremath{\pm}\scalebox{0.7}{0.89} & 5.85\ensuremath{\pm}\scalebox{0.7}{0.55} & 5.71\ensuremath{\pm}\scalebox{0.7}{0.38\phantom{0}} & 4.55\ensuremath{\pm}\scalebox{0.7}{3.06} & 25.97\ensuremath{\pm}\scalebox{0.7}{15.84} & 5.54\ensuremath{\pm}\scalebox{0.7}{0.51} & 8.07\phantom{0} \\
          & GAT   & 3.38\ensuremath{\pm}\scalebox{0.7}{0.45} & 4.35\ensuremath{\pm}\scalebox{0.7}{0.82} & 4.93\ensuremath{\pm}\scalebox{0.7}{0.75} & 6.64\ensuremath{\pm}\scalebox{0.7}{0.71\phantom{0}} & 6.18\ensuremath{\pm}\scalebox{0.7}{5.83} & 15.86\ensuremath{\pm}\scalebox{0.7}{5.83\phantom{0}} & 8.10\ensuremath{\pm}\scalebox{0.7}{1.24} & 7.06\phantom{0} \\
          & BWGNN & 19.48\ensuremath{\pm}\scalebox{0.7}{2.20} & 4.02\ensuremath{\pm}\scalebox{0.7}{0.43} & 5.30\ensuremath{\pm}\scalebox{0.7}{0.87} & 28.37\ensuremath{\pm}\scalebox{0.7}{4.31\phantom{0}} & 5.71\ensuremath{\pm}\scalebox{0.7}{1.60} & 10.47\ensuremath{\pm}\scalebox{0.7}{0.05\phantom{0}} & 6.79\ensuremath{\pm}\scalebox{0.7}{0.03} & 11.45\phantom{0} \\
          & GHRN  & 9.21\ensuremath{\pm}\scalebox{0.7}{2.94} & 5.37\ensuremath{\pm}\scalebox{0.7}{0.14} & 5.23\ensuremath{\pm}\scalebox{0.7}{0.19} & 24.57\ensuremath{\pm}\scalebox{0.7}{3.60\phantom{0}} & 3.61\ensuremath{\pm}\scalebox{0.7}{0.68} & 8.35\ensuremath{\pm}\scalebox{0.7}{2.44\phantom{0}} & 11.18\ensuremath{\pm}\scalebox{0.7}{1.10} & 9.65\phantom{0} \\
    \midrule
    \multirow{3}[2]{*}{\textit{Unsupervised}} & AnomalyDAE & 11.67\ensuremath{\pm}\scalebox{0.7}{0.38} & 31.37\ensuremath{\pm}\scalebox{0.7}{1.09} & 32.65\ensuremath{\pm}\scalebox{0.7}{1.46} & 7.17\ensuremath{\pm}\scalebox{0.7}{0.04\phantom{0}} & 1.32\ensuremath{\pm}\scalebox{0.7}{0.01} & 6.10\ensuremath{\pm}\scalebox{0.7}{0.08\phantom{0}} & 9.26\ensuremath{\pm}\scalebox{0.7}{1.51} & 14.22\phantom{0} \\
          & CoLA  & 3.55\ensuremath{\pm}\scalebox{0.7}{0.12} & 4.76\ensuremath{\pm}\scalebox{0.7}{0.21} & 5.86\ensuremath{\pm}\scalebox{0.7}{0.34} & 6.11\ensuremath{\pm}\scalebox{0.7}{0.36\phantom{0}} & 3.41\ensuremath{\pm}\scalebox{0.7}{1.38} & 9.67\ensuremath{\pm}\scalebox{0.7}{0.50\phantom{0}} & 6.55\ensuremath{\pm}\scalebox{0.7}{0.44} & 5.70\phantom{0} \\
          & TAM   & 3.74\ensuremath{\pm}\scalebox{0.7}{0.01} & 4.15\ensuremath{\pm}\scalebox{0.7}{0.02} & 5.58\ensuremath{\pm}\scalebox{0.7}{0.08} & 5.66\ensuremath{\pm}\scalebox{0.7}{0.17\phantom{0}} & 5.37\ensuremath{\pm}\scalebox{0.7}{0.51} & 6.20\ensuremath{\pm}\scalebox{0.7}{0.02\phantom{0}} & 7.90\ensuremath{\pm}\scalebox{0.7}{0.05} & 5.51\phantom{0} \\
    \midrule
    \multirow{5}[2]{*}{\textit{Generalist}} & UNPrompt & 24.02\ensuremath{\pm}\scalebox{0.7}{1.38} & 9.47\ensuremath{\pm}\scalebox{0.7}{0.91} & 10.26\ensuremath{\pm}\scalebox{0.7}{0.52} & 35.29\ensuremath{\pm}\scalebox{0.7}{0.90\phantom{0}} & 3.44\ensuremath{\pm}\scalebox{0.7}{0.31} & 28.41\ensuremath{\pm}\scalebox{0.7}{2.25\phantom{0}} & \underline{12.28}\ensuremath{\pm}\scalebox{0.7}{1.63} & 17.59\phantom{0} \\
          & ARC (10-shot) & \underline{36.53}\ensuremath{\pm}\scalebox{0.7}{0.24} & 30.93\ensuremath{\pm}\scalebox{0.7}{4.24} & 39.23\ensuremath{\pm}\scalebox{0.7}{4.40} & \textbf{37.77}\ensuremath{\pm}\scalebox{0.7}{0.35\phantom{0}} & 4.92\ensuremath{\pm}\scalebox{0.7}{1.85} & 59.47\ensuremath{\pm}\scalebox{0.7}{2.38} & 7.98\ensuremath{\pm}\scalebox{0.7}{1.18} & 30.98\phantom{0} \\
          & AnomalyGFM & 3.10\ensuremath{\pm}\scalebox{0.7}{0.17} & 3.69\ensuremath{\pm}\scalebox{0.7}{0.20} & 4.36\ensuremath{\pm}\scalebox{0.7}{0.09} & 4.85\ensuremath{\pm}\scalebox{0.7}{0.40\phantom{0}} & 6.00\ensuremath{\pm}\scalebox{0.7}{0.52} & 9.05\ensuremath{\pm}\scalebox{0.7}{1.03\phantom{0}} & 9.18\ensuremath{\pm}\scalebox{0.7}{1.69} & 5.75\phantom{0} \\
          & IA-GGAD & 36.33\ensuremath{\pm}\scalebox{0.7}{0.16} & \textbf{46.96}\ensuremath{\pm}\scalebox{0.7}{1.16} & \underline{45.03}\ensuremath{\pm}\scalebox{0.7}{2.06} & \underline{36.64}\ensuremath{\pm}\scalebox{0.7}{0.66\phantom{0}} & \underline{8.65}\ensuremath{\pm}\scalebox{0.7}{1.74} & \underline{60.16}\ensuremath{\pm}\scalebox{0.7}{0.93\phantom{0}} & 6.90\ensuremath{\pm}\scalebox{0.7}{0.29} & \underline{34.38}\phantom{0} \\
          & \textbf{\modelname} & \textbf{37.17}\ensuremath{\pm}\scalebox{0.7}{0.07} & \underline{40.15}\ensuremath{\pm}\scalebox{0.7}{1.31} & \textbf{45.25}\ensuremath{\pm}\scalebox{0.7}{0.88} & 35.31\ensuremath{\pm}\scalebox{0.7}{0.42\phantom{0}} & \textbf{9.00}\ensuremath{\pm}\scalebox{0.7}{1.12} & \textbf{60.26}\ensuremath{\pm}\scalebox{0.7}{0.30\phantom{0}} & \textbf{31.55}\ensuremath{\pm}\scalebox{0.7}{4.84} & \textbf{36.96}\phantom{0} \\
    \bottomrule
    \end{tabular}%
  }
  \label{tab:mainap}%
\end{table*}

\section{Experiments}
\label{sec:exp}

In this section, we conduct comprehensive experiments to systematically evaluate our proposed \modelname~framework. 
The evaluation is designed to answer the following key research questions:
\begin{itemize}
    \item \textit{RQ1: Overall Performance.} How does \modelname~perform against a wide range of supervised, unsupervised, and state-of-the-art generalist baselines in the challenging zero-shot cross-domain setting?
    \item \textit{RQ2: Key Component Analysis.} What are the precise contributions of the core innovations in \modelname, namely the anomaly-aware multi-curvature feature alignment, the mixture of Riemannian experts scorer, and the memory-based dynamic router, to overall model performance?
    \item \textit{RQ3: Hyperparameter Sensitivity.} How sensitive is \modelname~to its key hyperparameters, such as the embedding dimension and the number of experts, assessing its robustness and practical applicability?
\end{itemize}

\subsection{Experimental Setup}

\subsubsection{Datasets}
To evaluate the generalization capability of our model, we conduct experiments on 11 widely used benchmark datasets spanning citation, social, and co-review networks. Following a rigorous protocol, all models are trained on four source graphs (PubMed~\cite{sen2008collective}, Flickr~\cite{ding2019deep,tang2009relational}, Reddit~\cite{kumar2019predicting}, YelpChi~\cite{rayana2015collective,mcauley2013amateurs}) and evaluated on seven unseen target graphs (ACM~\cite{tang2008arnetminer}, Amazon~\cite{rayana2015collective,mcauley2013amateurs}, BlogCatalog~\cite{ding2019deep,tang2009relational}, Citeseer~\cite{sen2008collective}, Cora~\cite{sen2008collective}, Facebook~\cite{xu2022contrastive}, Weibo~\cite{kumar2019predicting}). Dataset statistics are summarized in Table~\ref{tab:datasets}.

\subsubsection{Baselines}
To ensure a rigorous and fair experimental validation, we benchmark the proposed \modelname~against an extensive collection of competitive methods:
\begin{itemize}
    \item \textit{Supervised Baselines:} These methods are trained on the source graphs with full label access. We include \textit{GCN}~\cite{kipf2016semi}, a foundational model that aggregates neighborhood features; \textit{GAT}~\cite{velivckovic2017graph}, which uses attention mechanisms for weighted neighbor aggregation; \textit{BWGNN}~\cite{tang2022rethinking}, which applies band-pass filters to amplify high-frequency signals associated with anomalies; and \textit{GHRN}~\cite{gao2023addressing}, which is designed to handle heterophily in graphs.
    
    \item \textit{Unsupervised Baselines:} These methods are trained on the source graphs without any labels. We include \textit{AnomalyDAE}~\cite{fan2020anomalydae}, a classic reconstruction-based method using a dual autoencoder for structure and attributes; \textit{CoLA}~\cite{liu2021anomaly}, a contrastive learning framework that learns representations by distinguishing nodes from their corrupted counterparts; and \textit{TAM}~\cite{qiao2023truncated}, which optimizes node embeddings based on a local affinity assumption.
    
    \item \textit{Generalist Baselines:} We include \textit{UNPrompt}~\cite{niu2024zero}, \textit{AnomalyGFM}~\cite{qiao2025anomalygfm}, and \textit{IA-GGAD}~\cite{zhangia} as zero-shot cross-domain baselines. We also report \textit{ARC}~\cite{liu2024arc} under a 10-shot target-label setting as a label-assisted reference, marked as ARC (10-shot) in the tables.
\end{itemize}

\subsubsection{Evaluation Metrics}

To evaluate anomaly detection performance comprehensively, we adopt two standard metrics: AUROC and AUPRC, which respectively assess overall discriminative ability and performance on the minority class.
To ensure the stability of our findings, we conduct 5 independent runs for every experiment using various random seeds, providing both average results and their corresponding standard deviations.

\subsubsection{Implementation Details}

For the detailed hyperparameter settings, the unified feature dimension $D$ after alignment is set to 32. The GNN backbone for structure encoding aggregates information from a 2-hop neighborhood, with residual connections used to construct final node embeddings.
For the MoE scorer, the number of experts $K$ is set to 5, and we use top-$k$ routing with $k=2$. The Riemannian experts are initialized with curvatures $[0.0,-0.5,-1.0,0.5,1.0]$, covering Euclidean, hyperbolic, and spherical geometries. The expert curvatures are learnable during training, allowing each expert to adapt its geometry to the source-graph distributions. The gating mechanism's softmax temperature is set to 0.7. We use Adam with a learning rate of $5e$-$5$ and a weight decay of $5e$-$5$, training for 40 epochs on the combined source graphs. The loss weights in Eq.~\eqref{eq:loss_total} are set as $\lambda_1=1.0$ ($\mathcal{L}_{\mathrm{embed}}$), $\lambda_2=0.5$ ($\mathcal{L}_{\mathrm{feat}}$), $\lambda_3=0.1$ ($\mathcal{L}_{\mathrm{struct}}$), $\lambda_4=0.1$ ($\mathcal{L}_{\mathrm{con}}$), and $\lambda_5=0.01$ ($\mathcal{L}_{\mathrm{gate}}$). The same hyperparameter configuration is used for all target datasets, without target-domain validation or fine-tuning. For all baselines, we adopt the optimal hyperparameter settings reported in their original papers or tune them on the source graphs. All experiments are conducted on a single NVIDIA Tesla A100 with 40GB of memory.

\subsection{Overall Performance (RQ1)}

The results in Table~\ref{tab:mainauc} and Table~\ref{tab:mainap} demonstrate the superiority of \modelname. Across seven target datasets, \modelname~achieves the best average AUROC of 82.09\% and AUPRC of 36.96\%, outperforming supervised, unsupervised, and zero-shot generalist baselines. Compared with the strongest zero-shot competitor IA-GGAD, \modelname~improves the average AUROC by 5.09\%. Notably, \modelname~also surpasses ARC (10-shot), which uses limited target labels, by 6.81\% AUROC and 5.98\% AUPRC on average. These results validate the benefit of explicitly modeling geometric heterogeneity for cross-domain GAD.

\subsection{Key Component Analysis (RQ2)}
\begin{figure*}[htbp] 
  \centering
  \includegraphics[width=\linewidth]{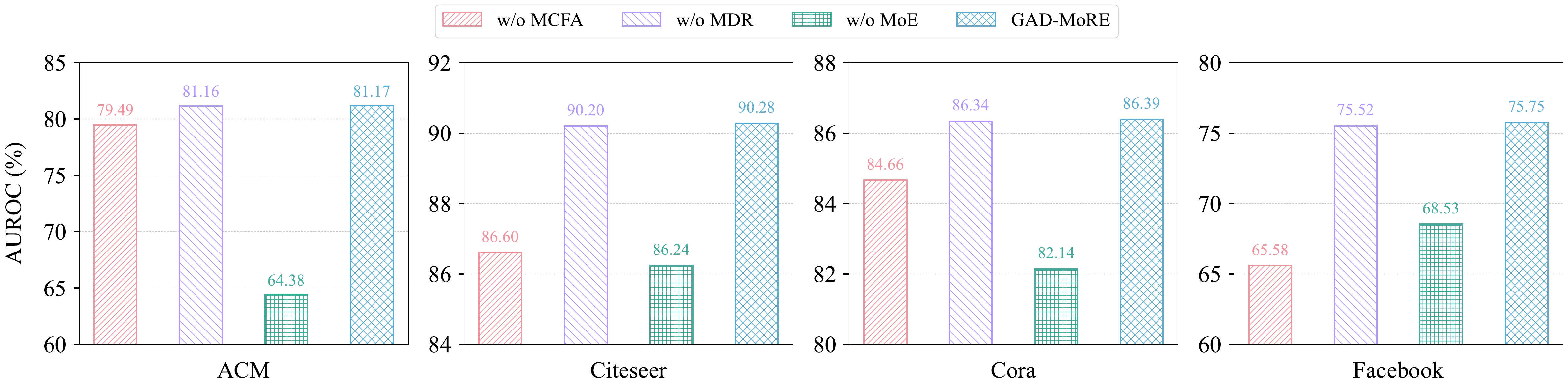} 
  \caption{AUROC performance of \modelname~and its ablation variants on target GAD datasets. All models are trained exclusively on four source graphs (PubMed, Flickr, Reddit, YelpChi) and evaluated in a zero-shot cross-domain setting.}
  \label{fig:ablation}
\end{figure*}

To answer our second research question and quantify the precise contributions of the central innovations within \modelname, we conduct a thorough ablation study. We design several variants of our model, each disabling or replacing a key component, and evaluate their AUROC performance against the full model on representative datasets. We evaluate the contribution of each core component by removing or replacing it individually. The resulting variants, whose acronyms match those reported in Figure~\ref{fig:ablation}, are defined as follows for a direct comparison with the full model:
\begin{itemize}
    \item \textit{w/o MCFA:} In this variant, we bypass the MCFA module and use a standard PCA for dimensionality reduction, removing the multi-geometry signals.
    \item \textit{w/o MoE:} 
    We replace the MoE scorer with an attention-based reconstruction scorer over reference embeddings.
    \item \textit{w/o MDR:} In this version, we replace our Memory-based Dynamic Router with a simpler linear router.
\end{itemize}

Figure~\ref{fig:ablation} presents the results of our ablation study. These findings clearly confirm the critical importance of each model component. The most significant performance degradation is observed in the w/o MoE variant, with the average AUROC dropping to 75.32\%. This provides the strongest empirical evidence for our central hypothesis: a single geometric lens is insufficient for generalizable GAD, and a collection of specialized Riemannian experts is essential for robust performance.
By contrast, removing the Multi-curvature Feature Alignment (w/o MCFA) also leads to a substantial performance drop, particularly on the Facebook dataset. This demonstrates that enriching initial node representations with signals from diverse Riemannian spaces provides a more robust foundation, enabling the model to adapt more effectively to unseen graph structures.
Lastly, replacing the Memory-based Dynamic Router (w/o MDR) with a simpler linear router results in a slight but consistent performance decrease across most datasets. This highlights the effectiveness of our proposed dynamic routing strategy, as history-aware assignments from the MDR lead to more precise final reconstruction.

\begin{figure*}[t]
    \centering
    \includegraphics[width=\linewidth]{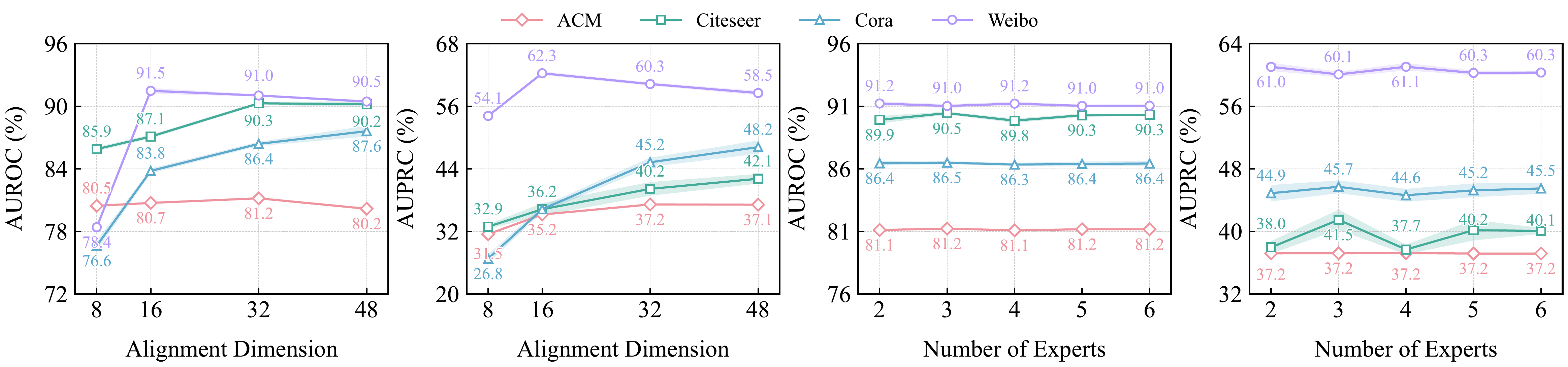}
    \caption{Hyperparameter sensitivity analysis of \modelname. The left two subplots show the impact of the unified embedding dimension $D$ on AUROC and AUPRC, while the right two subplots illustrate the sensitivity to the number of experts $K$.}
    \label{fig:sensitivity}
\end{figure*}

\subsection{Hyperparameter Sensitivity (RQ3)}

In this section, we investigate the sensitivity of \modelname~to its two most critical hyperparameters: the unified embedding dimension $D$ and the number of experts $K$. This analysis helps to understand the model's robustness and guides practical application. Figure~\ref{fig:sensitivity} illustrates the performance of \modelname~with respect to these two parameters on four representative test datasets, including ACM, Citeseer, Cora, and Weibo.

For the embedding dimension $D$ (left column of Figure~\ref{fig:sensitivity}), we observe that performance for both AUROC and AUPRC generally improves as the dimension increases from 16 to 32. This indicates that a larger dimension allows for more expressive representations to capture complex patterns. However, the performance then stabilizes or slightly decreases for dimension 48, suggesting that overly large dimensions may lead to overfitting on the source datasets without improving generalization. Our chosen value of $D=32$ offers the best trade-off between performance and model efficiency.

For the number of experts $K$ (right column of Figure~\ref{fig:sensitivity}), performance steadily increases as we add more experts from $K=2$ up to $K=5$. This suggests that a richer set of geometric lenses is beneficial for handling the diverse graph structures encountered in the test sets. Beyond $K=5$, the performance plateaus, indicating that five experts are sufficient to cover the geometric diversity in our datasets without becoming redundant or introducing unnecessary complexity. This confirms our choice of $K=5$ as optimal for the MoE scorer. Overall, the results show that \modelname~is robust to variations in these hyperparameters within a reasonable range.

\section{Conclusion}
\label{sec:conclusion}

In this paper, we introduce \modelname, a novel framework for generalizable graph anomaly detection that addresses the challenge of geometric heterogeneity. 
Motivated by our empirical finding that no single curvature is universally optimal, \modelname{} models diverse anomaly patterns with an anomaly-aware multi-curvature feature alignment module, a mixture of Riemannian experts scorer, and a memory-based dynamic router. 
The feature alignment module projects raw features into parallel Riemannian spaces to bridge geometry-agnostic attributes and manifold representations. 
The Riemannian experts then model anomaly patterns in distinct curvature spaces, while the memory-based router aligns expert selection with historical reconstruction feedback. 
Extensive experiments demonstrate that \modelname{} achieves state-of-the-art performance in zero-shot settings and consistently outperforms strong generalist competitors.

\section*{Acknowledgment}
The corresponding author is Qingyun Sun. This work is supported by NSFC under grants No.62427808 and No.U2541212, and by the Fundamental Research Funds for the Central Universities.

\bibliographystyle{IEEEtran}
\bibliography{ref}

\end{document}